\documentclass[letterpaper, 10 pt, journal, twoside]{IEEEtran}
\ifCLASSINFOpdf
\else
\fi

\usepackage{balance}
\usepackage{graphicx}
\usepackage{amsmath}
\usepackage{algorithmic}
\usepackage{array}
\usepackage[caption=false,font=normalsize,labelfont=sf,textfont=sf]{subfig}
\usepackage{fixltx2e}
\usepackage{dblfloatfix}
\usepackage{url}
\usepackage{colortbl}
\usepackage{booktabs}
\usepackage{multirow}
\usepackage{subfig}
\usepackage{svg}
\usepackage{float}
\usepackage{bm}
\makeatletter
\let\NAT@parse\undefined
\makeatother
\usepackage[colorlinks=true, linkcolor=blue, citecolor=blue, urlcolor=blue]{hyperref}
\newcommand{\ie}{\textit{i.e. }}

\begin{document}
%
\title{SLC$^2$-SLAM: Semantic-guided Loop Closure using Shared Latent Code for NeRF SLAM}
\author{Yuhang Ming$^1$, Di Ma$^1$, Weichen Dai$^1$, Han Yang$^1$, Rui Fan$^2$, Guofeng Zhang$^3$ and Wanzeng Kong$^{1\S}$
\thanks{Manuscript received: February 22, 2025; Accepted: March 16, 2025.}
\thanks{This paper was recommended for publication by Editor Javier Civera upon evaluation of the Associate Editor and Reviewers' comments.
This work was supported in part by National Natural Science Foundation of China under Grant 62401188, 62473288, the Fundamental Research Funds for the Central Universities, NIO University Programme (NIO UP), and the Xiaomi Young Talents Program.
} 

\thanks{$^{1}$Yuhang Ming, Di Ma, Weichen Dai and Han Yang are with School of Computer Science and Key Laboratory of Brain Machine Collaborative Intelligence of Zhejiang Province, Hangzhou Dianzi University, Hangzhou, 310018, China.
        {\tt\footnotesize \{yuhang.ming, 231050008, weichendai, yhan\_hdu, kongwanzeng\}@hdu.edu.edu}}%
\thanks{$^{2}$ Rui Fan is with the College of Electronics \& Information Engineering, Shanghai Institute of Intelligent Science and Technology, Shanghai Research Institute for Intelligent Autonomous Systems, the State Key Laboratory of Intelligent Autonomous Systems, and Frontiers Science Center for Intelligent Autonomous Systems, Tongji University, Shanghai 201804, China.
        {\tt\footnotesize rui.fan@ieee.org}}%
\thanks{$^{3} $Guofeng Zhang is with State Key Lab of CAD\&CG, Zhejiang University, Hangzhou, 310058, China.{\tt\footnotesize zhangguofeng@zju.edu.cn}}%
        
\thanks{Digital Object Identifier (DOI): see top of this page.}
}

\markboth{IEEE Robotics and Automation Letters. Preprint Version. Accepted MARCH, 2025}
{Ming \MakeLowercase{\textit{et al.}}: SLC$^2$-SLAM: Semantic-guided Loop Closure using Shared Latent Code for NeRF SLAM}

\maketitle

\begin{abstract}

Targeting the notorious cumulative drift errors in NeRF SLAM, we propose a Semantic-guided Loop Closure using Shared Latent Code, dubbed SLC$^2$-SLAM.
We argue that latent codes stored in many NeRF SLAM systems are not fully exploited, as they are only used for better reconstruction. 
In this paper, we propose a simple yet effective way to detect potential loops using the same latent codes as local features. 
To further improve the loop detection performance, we use the semantic information, which are also decoded from the same latent codes to guide the aggregation of local features.
Finally, with the potential loops detected, we close them with a graph optimization followed by bundle adjustment to refine both the estimated poses and the reconstructed scene.
To evaluate the performance of our SLC$^2$-SLAM, we conduct extensive experiments on Replica and ScanNet datasets. 
Our proposed semantic-guided loop closure significantly outperforms the pre-trained NetVLAD and ORB combined with Bag-of-Words, which are used in all the other NeRF SLAM with loop closure. 
As a result, our SLC$^2$-SLAM also demonstrated better tracking and reconstruction performance, especially in larger scenes with more loops, like ScanNet.
\end{abstract}

%
\IEEEpeerreviewmaketitle

\section{Introduction}
%
%
%
%

\begin{figure}[t]
\centering
\includegraphics[width=\linewidth]{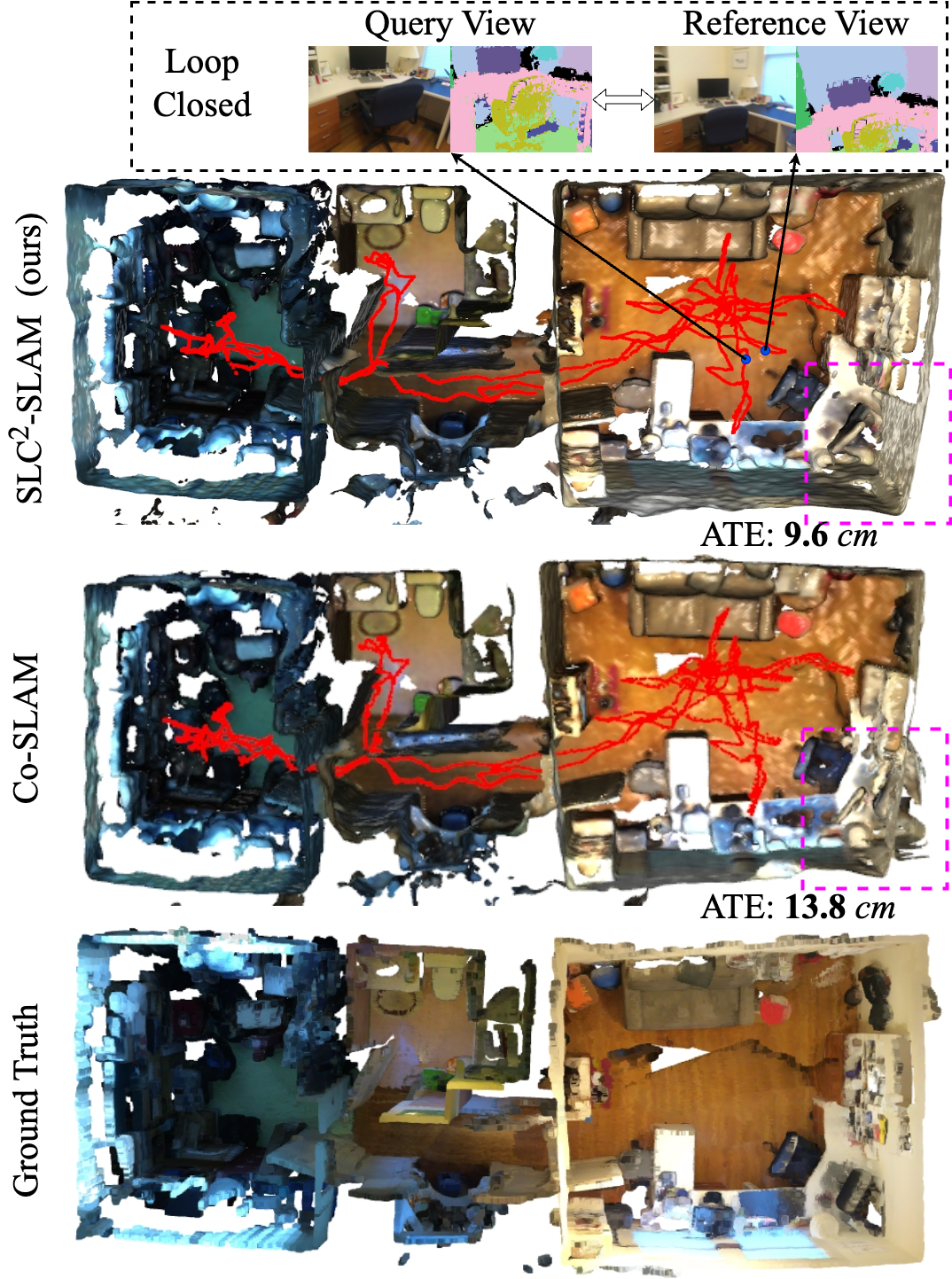}
\caption{Tracking and reconstruction results on the scene0054 of ScanNet~\cite{scannet}. 
With semantic-guided loop closure, our SLC$^2$-SLAM achieved better tracking and reconstruction performance. In contrast, our base system Co-SLAM~\cite{coslam} exhibits obvious misalignment, especially evident in the areas in the pink bounding boxes.
} 
\label{fig:teaser}
\vspace{-4ex}
\end{figure}

\IEEEPARstart{U}{sing} a RGB-D camera as the primary sensor, dense simultaneous localization and mapping (SLAM) aims at estimating the self-motion, \ie poses, of an agent while recovering the dense 3D reconstruction of its surrounding environment. Dense SLAM is the core to a wide range of spatial artificial intelligence (AI) applications, including autonomous robots and systems, embodied AI, and metaverse applications. Thus, it has been {a popular} research area in the robotics and computer vision communities.

Over the past decade, the field has seen remarkable advancements in dense SLAM, alongside a growing integration of SLAM systems with neural networks. 
Early dense SLAM systems, such as KinectFusion~\cite{kinectfusion} and ElasticFusion~\cite{elasticfusion}, prioritized precise geometrical reconstructions of environments, enabling detailed spatial modeling. 
Then, incorporating pre-trained neural networks, dense SLAM systems have evolved to provide enhanced scene comprehension~\cite{objectreloc} and increased resilience against cumulative drift errors~\cite{fdslam}. 
This synergy has expanded the scope of SLAM, transforming it from purely geometric mapping to a more semantically aware, robust system capable of more accurate and stable performance in complex environments.

More recently, the introduction of neural radiance fields (NeRF)~\cite{nerf} has showcased the powerful scene representation capabilities of multi-layer perceptrons (MLP). By encoding 3D scenes implicitly within the weights of an MLP, it generates compact neural implicit maps, which not only reduce the storage requirements of large-scale reconstructed scenes but also allow for efficient bundle adjustment of both estimated poses and the reconstructed map. Due to these advantages, NeRF has garnered substantial interest for developing dense SLAM systems that leverage neural implicit representations~\cite{nerf-survey}. Pioneered by iMAP~\cite{imap} and NICE-SLAM~\cite{niceslam}, a series of NeRF SLAM systems have emerged, showing notable advances in reconstruction quality~\cite{voxfusion}, tracking precision~\cite{idfslam}, and overall system efficiency~\cite{coslam}. 
These developments represent a promising shift toward more accurate, storage-efficient, and computationally feasible dense SLAM solutions.

Comparatively, much less attention has been paid to address the cumulative drift errors in NeRF SLAM systems. 
Existing implementations of loop closure in NeRF SLAM generally follow one of three main approaches: (1) employing handcrafted local features with global descriptor aggregation, such as ORB features~\cite{orb} paired with Bag-of-Words (BoW) descriptors~\cite{bow}; (2) utilizing pre-trained place recognition models, like NetVLAD~\cite{netvlad}; and (3) applying a simple covisibility score-based method. 
However, none of these techniques offer an optimal solution for NeRF SLAM. Covisibility score-based methods are too simple to close the loops with large drifts, while the other approaches require additional efforts for local feature extraction. These added steps not only increase computational overhead but also risk losing relevant information unique to NeRF representations, highlighting the need for a more specialized loop closure strategy tailored to the NeRF SLAM framework.

Recognizing that recent NeRF SLAM systems commonly utilize InstantNGP-style~\cite{instantngp} mapping for efficiency, where latent codes are learned on-the-fly and stored throughout operation, we observe that these latent codes' potential as local features for loop detection has been underutilized.
In this paper, we introduce Semantic-guided Loop Closure using Shared Latent Code for NeRF SLAM (SLC$^2$-SLAM), a simple yet effective approach specifically designed to leverage these latent codes for effective loop detection within NeRF SLAM systems.
In particular, our method uniquely repurposes these latent codes, initially intended solely for scene reconstruction, as local geometric features which are then aggregated into a global descriptor. To enhance this aggregation process, we incorporate semantic information, also decoded from the latent codes, guiding the selection of local latent codes for better aggregation. After identifying potential loops, 
we close the loop with a pose graph optimization, followed by bundle adjustment, to refine both the estimated pose and the reconstructed map.

We rigorously evaluate the performance of our SLC$^2$-SLAM with extensive experiments on Replica~\cite{replica} and ScanNet~\cite{scannet} datasets. Our method shows significant improvement in loop detection capabilities, achieving an average recall rate of 0.662---outperforming the closest competing approach, which achieves only 0.277. 
This enhanced loop detection also contributes to superior tracking accuracy and reconstruction quality, particularly evident in the larger scenes from the ScanNet dataset, as illustrated in Fig.~\ref{fig:teaser}. 

Our main contributions are summarized as follows:

\begin{itemize}
    \item To the best of our knowledge, we are the first to exploit the latent codes stored in many NeRF SLAM system not only for scene reconstruction but also for semantic segmentation and loop detection.
    

    \item We conduct extensive experiments on publicly available datasets and our SLC$^2$-SLAM consistently outperforms existing methods, achieving state-of-the-art performance in loop detection, reconstruction quality, and competitive performance in tracking accuracy.
\end{itemize}

\section{Related Work}
\subsection{NeRF SLAM with Latent Codes}
To enhance the reconstruction quality of NeRF SLAM systems, many approaches leverage latent codes to capture local scene structures, reducing the burden on the MLP for detailed map representation. While various terms such as features, embeddings, or latent codes are used across the literature, we refer to them collectively as latent codes here for consistency.

Vox-Fusion~\cite{voxfusion} pioneered the integration of neural implicit maps with explicit voxel structures by attaching on-the-fly learned latent codes to voxel vertices and utilizing an octree for efficient voxel indexing. 
Similar concepts also appear in systems like Co-SLAM~\cite{coslam}, ESLAM~\cite{eslam}, and VPE-SLAM~\cite{vpeslam}. 
Both Co-SLAM and VPE-SLAM followed voxel representations, but with distinct encoding design. Co-SLAM~\cite{coslam} builds on the InstantNGP~\cite{instantngp} framework, introducing a joint coordinate and parametric encoding with multi-resolution hashing and One-blob encoding. VPE-SLAM, alternatively, presents a voxel-permutohedral encoding that merges sparse voxels with multi-resolution permutohedral tetrahedral.
Contrasting with voxel-centric approaches, ESLAM~\cite{eslam} specifically favors a plane-based representation to retain latent codes. 

Building on this hybrid map representation, various works have been published to improve the systems' performance from different {aspects}. 
For a richer scene understanding, both SNI-SLAM~\cite{snislam} and NIS-SLAM~\cite{nisslam} expand NeRF SLAM by generating semantic maps, allowing for detailed scene labeling.
Regarding the accuracy of the reconstructed geometry, Hu et al.~\cite{hu2024} address issues related to incomplete depth data by introducing attentive depth fusion priors into the volume rendering process.
In terms of robustness, HERO-SLAM~\cite{heroslam} tackles abrupt viewpoint changes by implementing a hybrid enhanced robust optimization, while RoDyn-SLAM~\cite{rodynslam} improves dynamic object handling by removing dynamic rays from the reconstruction process with motion masks generated from optical flow and semantic information.

\begin{figure*}[t]
\centering
\vspace{1ex}
\includegraphics[width=0.95\textwidth]{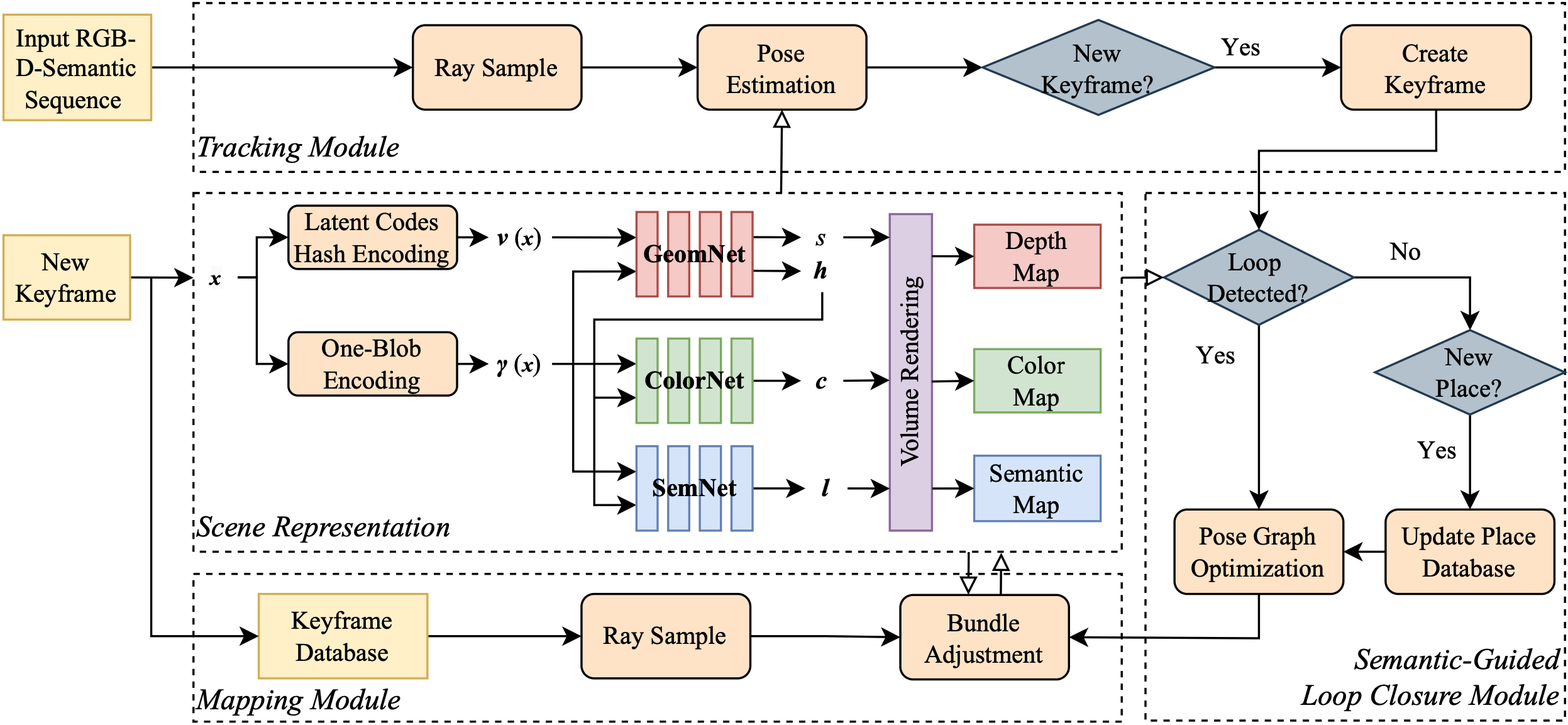}
\vspace{-1ex}
\caption{System Overview. Our proposed SLC$^2$-SLAM consists of four main components. At its core, there is a hybrid scene representation with latent code voxel hashing and three MLPs. Then, we have tracking module, mapping module, and semantic-guided loop closure module that interact with the hybrid scene representation to perform tracking, mapping, and loop closure.} 
\label{fig:overview}
\vspace{-3ex}
\end{figure*}

\subsection{NeRF SLAM with Loop Closures}
Loop closures are necessary to all SLAM systems to ensure robust operation in larger-scale environments. 
As outlined in the previous section, current loop closure approaches typically use one of three strategies: (1) covisibility scores, (2) the pre-trained NetVLAD~\cite{netvlad} model, or (3) ORB~\cite{orb} features in combination with BoW~\cite{bow} descriptors.
Additionally, NeRF SLAM systems can be broadly categorized by their approach to camera pose estimation: \textit{Coupled} NeRF SLAM, which estimates poses directly through inverse NeRF optimization, and \textit{Decoupled} NeRF SLAM, which leverages external SLAM systems for tracking. 

For \textit{Decoupled} NeRF SLAM systems, their choice of loop closure heavily rely on the type of tracker utilized. 
When employing DROID-SLAM~\cite{droidslam}, as seen in systems like Go-SLAM~\cite{goslam} and HI-SLAM~\cite{hislam}, the covisibility score---derived from the mean rigid flow---becomes the preferred option for loop detection due to its compatibility with DROID-SLAM's tracking mechanism. Alternatively, ORB-SLAM~\cite{orbslam2, orbslam3} is also widely used, featuring in systems such as Orbeez-SLAM~\cite{orbeezslam}, NEWTON~\cite{newton}, NGEL-SLAM~\cite{ngelslam}, and the system by Bruns \textit{et al.}~\cite{bruns2024}. These systems inherit ORB-SLAM’s loop closure capabilities, relying on ORB~\cite{orb} features paired with BoW~\cite{bow} descriptors for robust loop detection. Despite their strong tracking performance, these systems frequently encounter challenges in achieving high-quality reconstructions, as their focus on loop closure methods does not fully address limitations in fine-grained scene detail.

In \textit{Coupled} NeRF SLAM systems, where tracking and reconstruction are tightly integrated, various approaches have been explored for loop closure.
For instance, MIPS-Fusion~\cite{mipsfusion} introduced multi-implicit-submaps and performed submap-level loop closure by computing the covisibility between current frame and inactive submaps. However, this loop detection approach has limitations, primarily being effective for correcting only small drifts.
Vox-Fusion++\cite{voxfusionpp} instead relied on a pre-trained NetVLAD\cite{netvlad} model for loop detection and implemented a hierarchical pose optimization for robust loop closure. 
Similarly, Gaussian splatting SLAM systems such as GLC-SLAM~\cite{glcslam} and LoopSplat~\cite{loopsplat} also employed NetVLAD for loop detection.
Another approach, Loopy-SLAM~\cite{loopyslam} favors the combination of ORB~\cite{orb} features and BoW~\cite{bow} descriptors for loop detection, despite that these features were not part of the tracking or reconstruction process.
It can be seen that all these systems {require additional feature extraction steps} to achieve loop closure.

We argue that existing NeRF SLAM systems have not fully leveraged the latent codes inherent in their maps. By focusing solely on using these codes for reconstruction, they overlook the valuable potential of these latent codes in aiding loop detection directly, an oversight that our proposed approach seeks to address.

\section{System Design}
As shown in Fig.~\ref{fig:overview}, our SLC$^2$-SLAM comprises 4 components. At its core, we use a hybrid \textit{scene representation} with voxel-centric latent codes and three shallow MLPs. Interacting with this scene representation, we have a \textit{tracking module} that estimates the 6 degree-of-free (DoF) poses of the input frame, a \textit{mapping module} that is in charge of the keyframe management and scene representation optimization, and a \textit{semantic-guided loop closure module} that detects loops by aggregating on-the-fly learned latent codes and closes them with pose graph optimization.

\subsection{Hybrid Scene Representation}
Although our SLC$^2$-SLAM is able to work with any NeRF SLAM systems with latent codes, as we have reviewed above, we base our system on the Co-SLAM~\cite{coslam}, modify it to incorporate semantic information, and carry out all the experiments.

Following Co-SLAM, we use a sparse set of voxels, that are indexed by a hash table, with a learnable compact latent code attached to each voxel center/vertices for coordinate encoding, and employ OneBlob encoding for parametric encoding. 
Regarding the shallow MLPs, our scene representation contains three: the GeomNet, ColorNet and SemNet. 
In particular, the GeomNet takes in the latent code $\bm{v}(\bm{x})$ and the parametrically encoded position $\gamma(\bm{x})$, and outputs the scene geometry $s$, in terms of signed distance function (SDF), and a hidden feature vector $\bm{h}$. 
Connected to the GeomNet, ColorNet and SemMet are placed in parallel, both of which take the hidden feature vector $\bm{h}$ and the parametrically encoded position $\gamma(\bm{x})$ as input, and produce the RGB color $\bm{c}$ and semantic label $l$ respectively.

Then, given the input RGB-D frames and semantic masks, both the latent codes and the weights of the three MLPs can be learned with the following loss function:
\begin{equation}
\begin{split}
    \mathcal{L} = & \lambda_{rgb} \mathcal{L}_{rgb} + \lambda_{d} \mathcal{L}_{d} + \lambda_{sdf} \mathcal{L}_{sdf} + \lambda_{fs} \mathcal{L}_{fs} + \\
    & \lambda_{sem} \mathcal{L}_{sem} + \lambda_{smooth} \mathcal{L}_{smooth},
\end{split}
\end{equation}
{where each $\lambda$ represents a per-loss weight. $\mathcal{L}_{rgb}$, $\mathcal{L}_{d}$, $\mathcal{L}_{sdf}$, $\mathcal{L}_{fs}$, and $\mathcal{L}_{smooth}$ denote the loss terms for color, depth, SDF, free space, and smooth regularization, respectively, following the formulation in Co-SLAM~\cite{coslam}. The remaining, $\mathcal{L}_{sem}$, computes the cross-entropy loss for the semantic labels as follow: }
\begin{equation}
 \mathcal{L}_{sem} =  -\frac{1}{N} \sum_{i=1}^{N} \sum_{j=1}^{C} l_{i,j} \log(\hat{l}_{i,j}).
\end{equation}

\subsection{Keyframe Management}
\label{sec::kf-management}
{To strike a balance between map update frequency, loop detection efficiency, and runtime performance,}
our SLC$^2$-SLAM introduces a hierarchical keyframe management strategy, consisting of keyframes, covisible frames, and place frames. 

For keyframes, we follow Co-SLAM~\cite{coslam} and add a new keyframe every 5 frames. 
This high rate of keyframe addition enables frequent map optimization iterations, ensuring high reconstruction quality.
However, this density of keyframes introduces redundancy, which can be inefficient for loop detection and pose graph optimization.
To address this, we introduce {place frames}, a sparser subset of frames within the keyframe set, to streamline the loop detection process and {enhance the efficiency of pose graph optimization}. 

{We determine whether a keyframe should be added as a new place frame based on point cloud overlap.} Specifically, we calculate the overlap between the point cloud from the most recent keyframe and those from all previously stored place frames. If the overlap is below a user-defined threshold, $\tau_{place}$, the keyframe is accepted as a new place frame.
In practice, setting $\tau_{place}$ to a relatively low value ensures minimal overlap, resulting in only a few place frames per indoor room, which efficiently covers the scene with a reduced number of frames. 

{In contrast to} the spatially dense keyframes, {we find that, in practice, place frames are too sparse to effectively distribute accumulative errors detected during loop closures}. To balance between these extremes, we further introduce covisible frames, which have an intermediate spatial density. This density is also managed by using point cloud overlap, but with a higher threshold, $\tau_{covis}$, than that of place frames. 
{Importantly, all place frames are also designated as covisible frames. When a loop is detected at a keyframe, the keyframe, along with all covisible frames, are used to construct the pose graph for optimization.} The optimization process will be discussed in detail in Sec.~\ref{sec:lc-optim}.


\subsection{Semantic-Guided Loop Detection}

To perform loop detection, we formulate it as a retrieval task and solve it with a two-step process: generating global descriptors from local features and matching these descriptors with those in a database of known locations.

Unlike previous loop detection methods relying on handcrafted point features~\cite{orb, loopyslam} or neural features extracted by pre-trained convolutional neural networks~\cite{netvlad, voxfusionpp},
we propose directly leveraging the latent codes stored within the NeRF SLAM map as local features.
Since these latent codes are shared across tracking, mapping, and semantic segmentation tasks, this approach not only removes the need for external feature extractors but also enhances the overall efficiency of the system.

\begin{figure}[t]
\centering
\vspace{0.5ex}
\includegraphics[width=\linewidth]{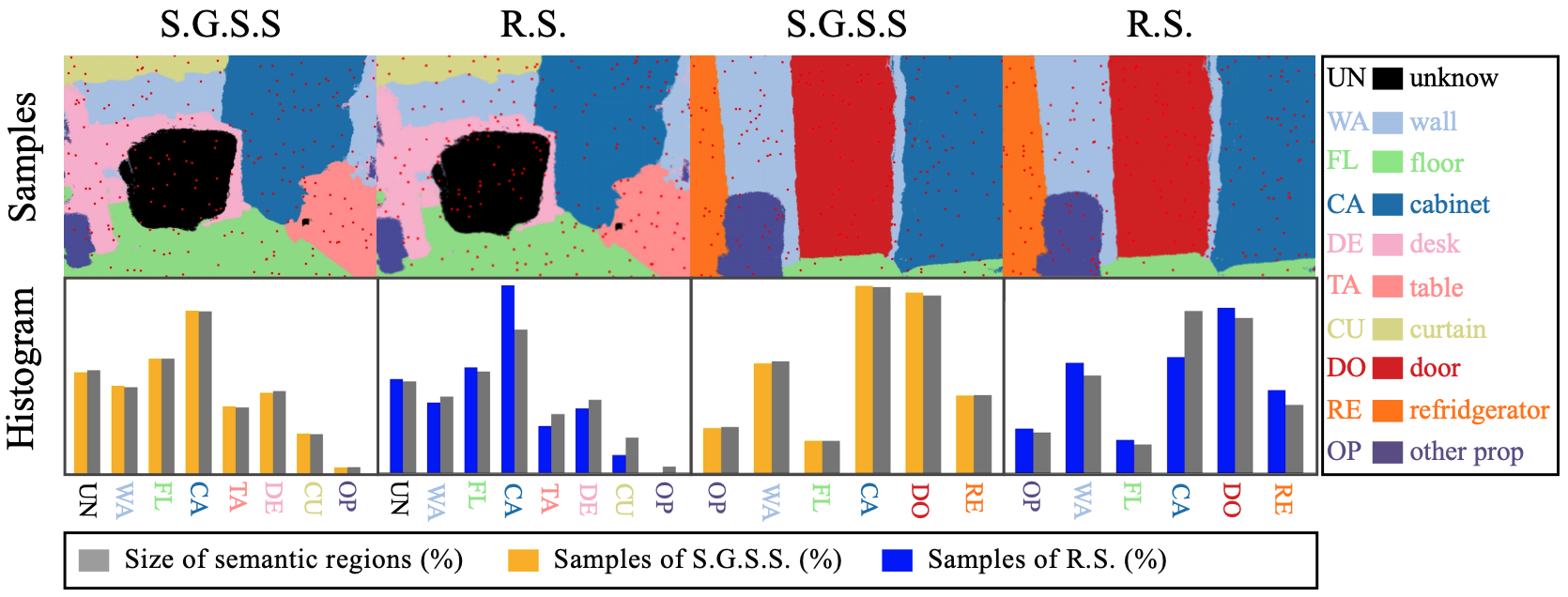}
\vspace{-4.5ex}
\caption{Examples of semantic-guided stratified sampling (S.G.S.S) versus random sampling (R.S.).}
\label{fig:sampling}
\vspace{-4ex}
\end{figure}

Given the high resolution of the input images, aggregating latent codes for every pixel is computationally {intractable}. Therefore, {we aim to select $M$ representative pixels that best describe the image.} To achieve this, we introduce a semantic-guided stratified sampling method {that utilizes the semantic masks predicted by SemNet and sets the number of samples to be proportional to the size of each semantic region.} 
Although naive random sampling could be used, Fig.~\ref{fig:sampling} illustrates that semantic-guided stratified sampling provides a more accurate view representation by 
{identifying the easily-overlooked small semantic regions and reducing oversampling of the dominant semantic region.}
The results in {Table~\ref{tab:Loop-scannet}} further demonstrate the advantages of our semantic-guided stratified sampling over random sampling.

After gathering latent codes to represent local features, we apply the vector of locally aggregated descriptors (VLAD)~\cite{vlad} to construct global descriptors for the current keyframe and all stored place frames. We then match these descriptors to identify the closest place frame, forming a loop hypothesis for the current keyframe.

To prevent catastrophic system failures caused by incorrect loop closures, we subject each loop hypothesis to further validation using both geometric and semantic information. Specifically, we calculate the overlap between the point clouds and their semantic labels. The loop hypothesis is only accepted if both overlaps exceed pre-set thresholds.

\subsection{Pose Graph Optimization}
\label{sec:lc-optim}

Given a pose graph comprising the covisible frames discussed in the previous section, the current keyframe as the loop frame, and its matched place frame, we can now incorporate a loop edge into the pose graph. This loop edge represents the relative transformation between the loop frame and its matched place frame, calculated using a standard point-to-plane iterative closest point (ICP)~\cite{ppicp} algorithm.

Once the graph is constructed, we proceed with optimization using standard pose graph toolkits. The recent release of PyPose~\cite{pypose} enables seamless integration of geometry-based optimization with learning-based loop detection, all within the PyTorch framework. Specifically, we employ the Levenberg-Marquardt optimizer, establishing a trust-region strategy to dynamically adjust the learning rate. For a particular edge in the pose graph, we define the loss function as follows:
\vspace{-1ex}
\begin{equation}
\mathbf{e}_i = \log(\mathbf{T}_{\text{edge}_i}^{-1} \mathbf{T}_{\text{node}_0} \mathbf{T}_{\text{node}_1}^{-1}),
\end{equation}
where $\mathbf{T}_{\text{edges}_i}$ is the relative transformation of between two frames connected by the edge, $\mathbf{T}_{\text{node0}}$ and$\mathbf{T}_{\text{node1}}$ are poses of the two frames respectively.
Then, the overall loss to be optimized can be formulated as:
\vspace{-1ex}
\begin{equation} 
\mathcal{L}_{pg} = \sum_{i} \|\mathbf{e}_i\|^2.
\end{equation}

After optimizing all covisible frames upon loop closure, we use them to update the poses of the keyframes. These updated keyframe poses are then included in an additional bundle adjustment step, {as in Co-SLAM~\cite{coslam}}, to jointly refine both the keyframe poses and the map, which is represented by the MLPs and latent codes.

\section{EXPERIMENTS}
\subsection{Experiment Setup}
We evaluate our proposed SLC$^2$-SLAM on two widely-used indoor datasets: Replica~\cite{replica} and ScanNet\textcolor{black}{v1}~\cite{scannet}. Replica is a synthetic dataset containing 18 high-fidelity replicates of different indoor rooms, offering ground-truth dense reconstruction, semantic and instance annotations, among other resources. In line with other NeRF SLAM studies, we use the subset provided in NICE-SLAM~\cite{niceslam}, comprising 2000-frame sequences from 8 out of the 18 indoor rooms. ScanNet, by contrast, is a large-scale dataset of 1,513 sequences collected from 707 real-world indoor rooms. Also to align with other NeRF SLAM papers, we use scenes \textit{0000, 0059, 0106, 0169, 0181}, and \textit{0207} for both qualitative and quantitative evaluations. 

To quantitatively assess the loop closure performance, we collect a database of place frames and a set of query frames from the sequences of the 6 ScanNet scenes.
In particular, all the place frames and query frames are the keyframes generated from the sequence at intervals of every 5 frames. Then, we use the overlap between frame-pairs as the criteria and select a keyframe as a place frame if the overlap is lower than 0.3. All other keyframes are used as query frame for evaluation. In total, from the 6 ScanNet scenes, we gathered 
96 
place frames and 
3,218 
query frames.

\textcolor{black}{
Pushing it to the limit, we further evaluated our method on the complete ScanNetv1~\cite{scannet} under the place recognition setup. Following the protocols established in recent indoor place recognition studies~\cite{cgis,aegis}, we utilized the test split of ScanNetv1 with spatial sparsification, resulting in 4,313 frames from 142 indoor rooms. 
From this set, we selected 294 frames as our place frames for retrieval by enforcing a minimum spatial separation of 3 meters, and used the remaining 4,019 frames as query frames.
It is important to note that, compared to the loop detection setup,
the place recognition setup uses significantly sparser place frames, making the task considerably more challenging.
}

\textbf{Evaluation Metrics:}
To evaluate tracking and reconstruction performance, we adopt standard evaluation metrics: root mean square {error (RMSE)} of the absolute trajectory error (ATE) for tracking, and accuracy (Acc.), completion (Comp.), and completion ratio (Comp. Ratio) for reconstruction. 
{Note that we follow Co-SLAM~\cite{coslam} to perform mesh culling before evaluation, and we refer readers to the survey paper~\cite{nerf-survey} for more details on metric computation.}
For loop detection, a loop candidate is accepted if the translational pose difference with the place frame is less than 1 meter and the rotational difference is under 35 degrees. We evaluate loop detection with three metrics---precision, recall, and F1 score---all based on top-1 retrieval results.
\textcolor{black}{As for place recognition, we follow the protocols used in \cite{cgis, aegis} and compute the average recall rates for Top-$K$ retrievals.} 

\textbf{Baselines:}
We choose Co-SLAM~\cite{coslam} as our primary baseline, as it serves as the foundation for our proposed SLC$^2$-SLAM. 
We then compare SLC$^2$-SLAM to three recent NeRF SLAM systems that support loop closure. 
Of particular interest are systems with a similar mapping setup, maintaining a single, global NeRF-based map. 
Therefore, we select Loopy-SLAM~\cite{loopyslam} and Orbeez-SLAM~\cite{orbeezslam} for comparison. 
To broaden the scope, we also include 
\textcolor{black}{two NeRF SLAM systems~\cite{niceslam, hu2024},
two loop closure-enabled NeRF SLAM systems that utilize submaps~\cite{bruns2024, mipsfusion}, a Gaussian splatting SLAM~\cite{SplaTAM}, and a loop closure-enabled Gaussian splatting SLAM~\cite{glcslam}.
} 

Regarding loop detection \textcolor{black}{and place recognition}, we compare our semantic-guided loop detection method with two commonly used approaches: NetVLAD~\cite{netvlad} and the combination of ORB~\cite{orb} and BoW~\cite{bow}.
Following the setup in GLC-SLAM~\cite{glcslam} and Loopy-SLAM~\cite{loopyslam}, we used the NetVLAD model pre-trained on the Pitts30K dataset~\cite{pitts} and the BoW vocabulary provided by ORB-SLAM2~\cite{orbslam2}. \textcolor{black}{Although not yet integrated in SLAM systems, two state-of-the-art (SoTA) Dinov2-based models, AnyLoc~\cite{anyloc} and SALAD~\cite{salad}, are also compared in the place recognition task to highlight the superiority of our method.}


\textbf{Implementation Details:}
We keep most of our system and training parameters in line with our backbone system, Co-SLAM~\cite{coslam}. For the additional modules, we design our SemNet as a 4-layer MLP with 32 hidden neurons, setting its learning rate 0.05, and assigning a weight of $\lambda_{sem}=10$ for the semantic loss.
Moreover, the thresholds for generating place frames and covisible frames are set to $\tau_{place}=0.3$, $\tau_{covis}=0.45$, respectively.
When a loop candidate is validated, we perform another 10 iterations of bundle adjustment following the pose graph optimization.
All of our experiments are performed on a desktop PC with AMD Ryzen 9 5950X CPU and NVIDIA GeForce RTX 4090 GPU.

\begin{table*}[t]
\vspace{1ex}
\caption{Loop detection results on ScanNet. The best and second best results are marked with \textbf{Bold} and \underline{Underline}.}
\vspace{-1ex}
\label{tab:Loop-scannet}
\resizebox{\textwidth}{!}{ 
  \begin{minipage}{\textwidth}  
    \centering
    \begin{tabular}{@{\extracolsep\fill}llcccccccc}
      \toprule
      Methods & Metric & scene0000 & scene0059 & scene0106 & scene0169 &scene0181 & scene0207 & Avg.  \\
      \midrule

        \multirow{3}{*}{NetVLAD~\cite{netvlad}} 
        & \textbf{Precision}$\uparrow$ 
        & 0.041 & 0.056 & 0.089 &0.089  &0.046 &  0.067 &0.063   \\
        & \textbf{F1-score}$\uparrow$ 
        & 0.028 & 0.083 &0.108 & 0.114 & 0.085    &  0.121&0.084  \\
        & \textbf{recall@1}$\uparrow$ 
        &0.022  & 0.158 &0.136  &0.158  &\underline{0.597}  &  \textbf{0.635}&0.277    \\
       \midrule

        \multirow{3}{*}{ORB~\cite{orb} + BoW~\cite{bow}} 
        & \textbf{Precision}$\uparrow$ 
        & 0.175 &\textbf{0.246}     & \textbf{0.645}& 0.298 & 0.273 & 0.335 &0.329 \\
        & \textbf{F1-score}$\uparrow$ 
        & 0.229 &\textbf{0.494}  & 0.142 & 0.330 &0.292 & 0.344  &0.305 \\
        & \textbf{recall@1}$\uparrow$ 
        & 0.331  & 0.167 & 0.080   &0.369 &0.314  &0.353 &0.269   \\
       \midrule
       
         \multirow{3}{*}{Ours (w/o semantic)} 
        & \textbf{Precision}$\uparrow$ 
        & \underline{0.306}   &0.147 &0.377 & \textbf{0.547}&\textbf{0.328} & \textbf{0.493} &\textbf{0.366} \\
        & \textbf{F1-score}$\uparrow$ 
        & \underline{0.336} &0.182  &\underline{0.462} &\textbf{0.605 }&\underline{0.352}& \underline{0.506}  &\underline{0.407}\\
        & \textbf{recall@1}$\uparrow$ 
        & \underline{0.371} &\underline{0.238} &\underline{0.597}   & \underline{0.677}  &0.378      & 0.520 &\underline{0.464}   \\
     \midrule
     
       \multirow{3}{*}{Ours} 
        & \textbf{Precision}$\uparrow$ 
        & \textbf{0.317}    &\underline{0.244} & \underline{0.379}& \underline{0.320} &\underline{0.286}  & \underline{0.461} &\underline{0.335} \\
        & \textbf{F1-score}$\uparrow$ 
        & \textbf{0.414} & \underline{0.325}  & \textbf{0.492}& \underline{0.462} &\textbf{0.416}&\textbf{0.521}&\textbf{0.438}  \\
        & \textbf{recall@1}$\uparrow$ 
        & \textbf{0.598}  &\textbf{0.489} &\textbf{0.701}    & \textbf{0.828}  &\textbf{0.756} & \underline{0.598} &\textbf{0.662} \\
     \bottomrule
    \end{tabular}

  \end{minipage}
}
\end{table*}

\begin{table}[t]
\vspace{-3ex}
\caption{\textcolor{black}{Average Recall Rate. The best and second best results are marked with \textbf{Bold} and \underline{Underline}.}}
\vspace{-1ex}
\label{tab:recall_comparison}
\centering
\begin{tabular}{lccc}
    \toprule
    \textcolor{black}{Methods} & \textcolor{black}{Recall@1} & \textcolor{black}{Recall@2} & \textcolor{black}{Recall@3} \\
    \midrule
    \textcolor{black}{NetVLAD~\cite{netvlad}}     & \textcolor{black}{0.113} & \textcolor{black}{0.164} & \textcolor{black}{0.206} \\
    \textcolor{black}{ORB~\cite{orb} + BoW~\cite{bow}}                & \textcolor{black}{0.165} & \textcolor{black}{0.195} & \textcolor{black}{0.220} \\
    \textcolor{black}{AnyLoc~\cite{anyloc}} & \textcolor{black}{0.208} & \textcolor{black}{0.284} & \textcolor{black}{0.327}  \\ 
    
  
    \textcolor{black}{SALAD~\cite{salad}}  & \textcolor{black}{\underline{0.322}} & \textcolor{black}{\underline{0.417}} & \textcolor{black}{\underline{0.462}} \\
    
    \textcolor{black}{Ours}                   & \textcolor{black}{\textbf{0.596}} & \textcolor{black}{\textbf{0.664}} & \textcolor{black}{\textbf{0.689}} \\
    \bottomrule
    \end{tabular}
    \vspace{-1ex}        
\end{table}

\begin{table}[t]

    \resizebox{\linewidth}{!}{%
    \begin{minipage}{1.1\linewidth}
    \centering
     \caption{Tracking results on ScanNet (ATE RMSE [cm]$\downarrow$). The best and second best results are marked with \textbf{Bold} and \underline{Underline}.}
     \vspace{-1ex}
   
    \begin{tabular}{lc c c c c c c}
        \toprule
         Methods & 0000 & 0059 & 0106 & 0169 & 0181 & 0207 & Avg. \\ 
         \midrule
        
        
        
         Co-SLAM~\cite{coslam} & 7.13 & 11.14 & 9.36 & \underline{5.90} & 11.81& 7.14 &8.75\\
        
         Loopy-SLAM~\cite{loopyslam}$^1$ & \textbf{4.2} & \underline{7.5} & 8.3 & 7.5  & \textbf{10.6} & 7.9 & \underline{7.7} \\  
        
         Orbeez-SLAM~\cite{orbeezslam} & 7.22 & \textbf{7.15} & {8.05} & 6.58  & 12.77 & 7.16 & 8.66 \\
        
         MIPS-Fusion~\cite{mipsfusion}$^1$ & 7.9 & 10.7 & 9.7  & 9.7  & 14.2 & 7.8 & 10.0 \\
         
         {SplaTAM~\cite{SplaTAM}} & {12.83} & {10.10} & {17.12} & {12.08} & {11.10} & {7.46} & {11.88} \\

         {GLC-SLAM~\cite{glcslam}$^1$} & {12.9} & {7.9} & {\textbf{6.3}} & {10.5} & {\underline{11.0}} & {\underline{6.3}} & {9.2}\\

         SLC$^2$-SLAM (ours) & \underline{5.83} & 9.45 & \underline{8.00} &\textbf{5.29}  & {11.26}   & \textbf{6.10}  & \textbf{7.66}\\ 
         \bottomrule

        \multicolumn{8}{l}{%
      \begin{minipage}{0.9\linewidth}%
        \vspace{1ex}
        \footnotesize 
        $^1$ Loopy-SLAM, MIPS-Fusion, and {GLC-SLAM} papers only reported tracking results in one decimal place.
      \end{minipage}%
    }\\
   
    \end{tabular}
    \label{tab:ate-scannnet}
    \end{minipage}
    }
\vspace{-3ex}
\end{table}

\begin{table*}
    \centering
    \caption{Tracking results on Replica (ATE RMSE [cm]$\downarrow$). The best and second best results are marked with \textbf{Bold} and \underline{Underline}}
    \begin{tabular}{lc c c c c c c c c}
        \toprule
        
        Methods & Room0 & Room1 & Room2  & Office0 & Office1 & Office2 & Office3 & Office4 & Avg.  \\ 
        \midrule
      
        NICE-SLAM~\cite{niceslam} & 1.69 & 2.04 & 1.55 & 0.99 & 0.90 & \underline{1.39} & 3.97 & 3.08 & 1.95 \\
       
        Co-SLAM~\cite{coslam}$^1$ & \underline{0.77} & \underline{1.04} & \underline{1.09} & \underline{0.58} & \underline{0.53} & 2.05 & \underline{1.49} & \underline{0.84} & \underline{0.99}  \\
        
        Hu \textit{et al.}~\cite{hu2024} & 1.39& 1.55 & 2.60& 1.09 & 1.23 & 1.61 & 1.61& 1.42 & 1.81 \\
       
        MIPS-Fusion~\cite{mipsfusion}$^2$ & 1.1 & 1.2 & 1.1 & 0.7 & 0.8 & \textbf{1.3} & 2.2& 1.1 &1.2  \\
        
        SLC$^2$-SLAM (ours) & \textbf{0.58} & \textbf{0.63} & \textbf{0.88} & \textbf{0.49}  & \textbf{0.49} & 1.53 & \textbf{1.37} & \textbf{0.66} & \textbf{0.83}\\ 
        \bottomrule

    \multicolumn{10}{l}{%
  \begin{minipage}{0.7\textwidth}%
    \vspace{1ex}
    \footnotesize 
    $^1$ The results for Co-SLAM are generated using their official implementation;\\
    $^2$ MIPS-Fusion paper only reported tracking results in one decimal place.
  \end{minipage}%
}\\
    \end{tabular}   
    \label{tab:ate-replica}
    \vspace{-2ex}
\end{table*}

\subsection{Results and Discussions}

\textbf{Loop Detection:} 
We present the quantitative results in Table~\ref{tab:Loop-scannet} \textcolor{black}{and \ref{tab:recall_comparison}}, all recorded prior to the loop validation step. Our semantic-guided approach shows a significant performance improvement over these baseline \textcolor{black}{and SoTA} methods across all scenes \textcolor{black}{and nearly all metrics in both loop detection and place recognition setups.}

We also conducted an ablation study by removing semantic guidance and substituting semantic-guided stratified sampling with naive random sampling. This modification led to a noticeable performance decline in recall and F1 score; however, our method still outperformed NetVLAD~\cite{netvlad} and the ORB~\cite{orb} and BoW~\cite{bow} combination.

\begin{figure}[t]
\centering
\vspace{-3ex}
\includegraphics[width=0.9\linewidth]{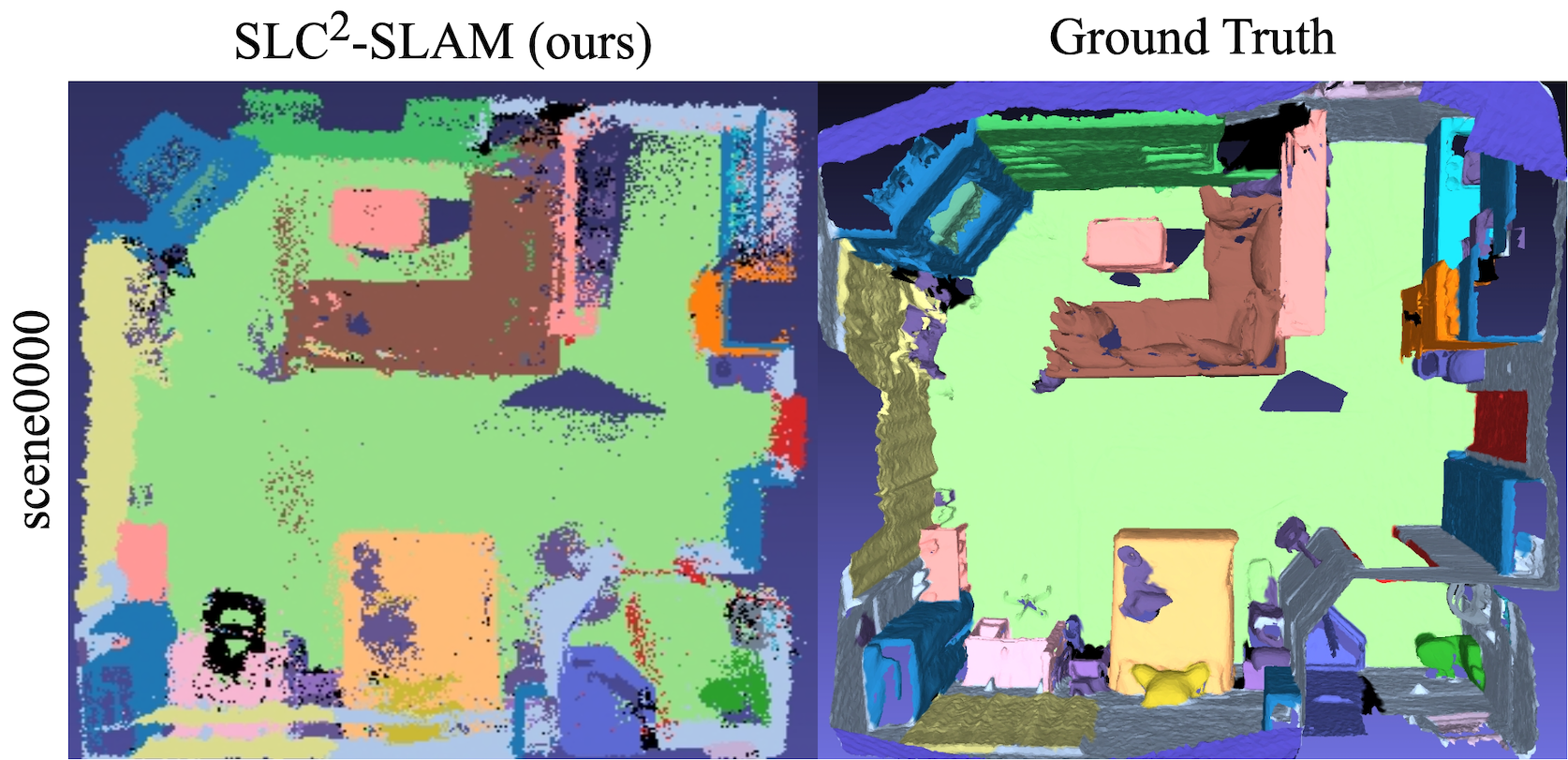}
\vspace{-2ex}
\caption{Semantic segmentation examples on ScanNet.
} 
\label{fig:sem}
\vspace{-3ex}
\end{figure}

In addition, though the semantic segmentation task is not the focus of our work but merely an assistant in the loop detection task, we provide some qualitative results shown in Fig.~\ref{fig:sem}. Quantitatively, our SLC$^2$-SLAM achieves a mean intersection over union (mIoU) of 0.6795 on the six test scenes of ScanNet~\cite{scannet}.
{In comparison, a recent semantic SLAM system, SGS-SLAM~\cite{sgsslam}, achieves 0.6980 on the same six scenes.
Although our system's performance is slightly lower, it is sufficient to effectively guide the loop detection process.}


\textbf{Tracking:} 
The tracking results, shown in Tables~\ref{tab:ate-replica} and~\ref{tab:ate-scannnet}, reveal that our approach outperforms the baseline system Co-SLAM~\cite{coslam}, with tracking accuracy gains of $16.16\%$ on Replica~\cite{replica} and $12.46\%$ on ScanNet~\cite{scannet}. These substantial improvements confirm the effectiveness of our loop closure method.
Additionally, compared to other recent systems, both with and without loop closure capabilities, our SLC$^2$-SLAM surpasses thess methods across most of the test scenes. Particularly in larger indoor rooms in ScanNet, our system shows superior average tracking performance. We attribute this to our system’s ability to detect more loops, enabling additional pose graph optimizations that enhance tracking accuracy.

\begin{figure*}[t]
    \centering
    \includegraphics[width=\textwidth]{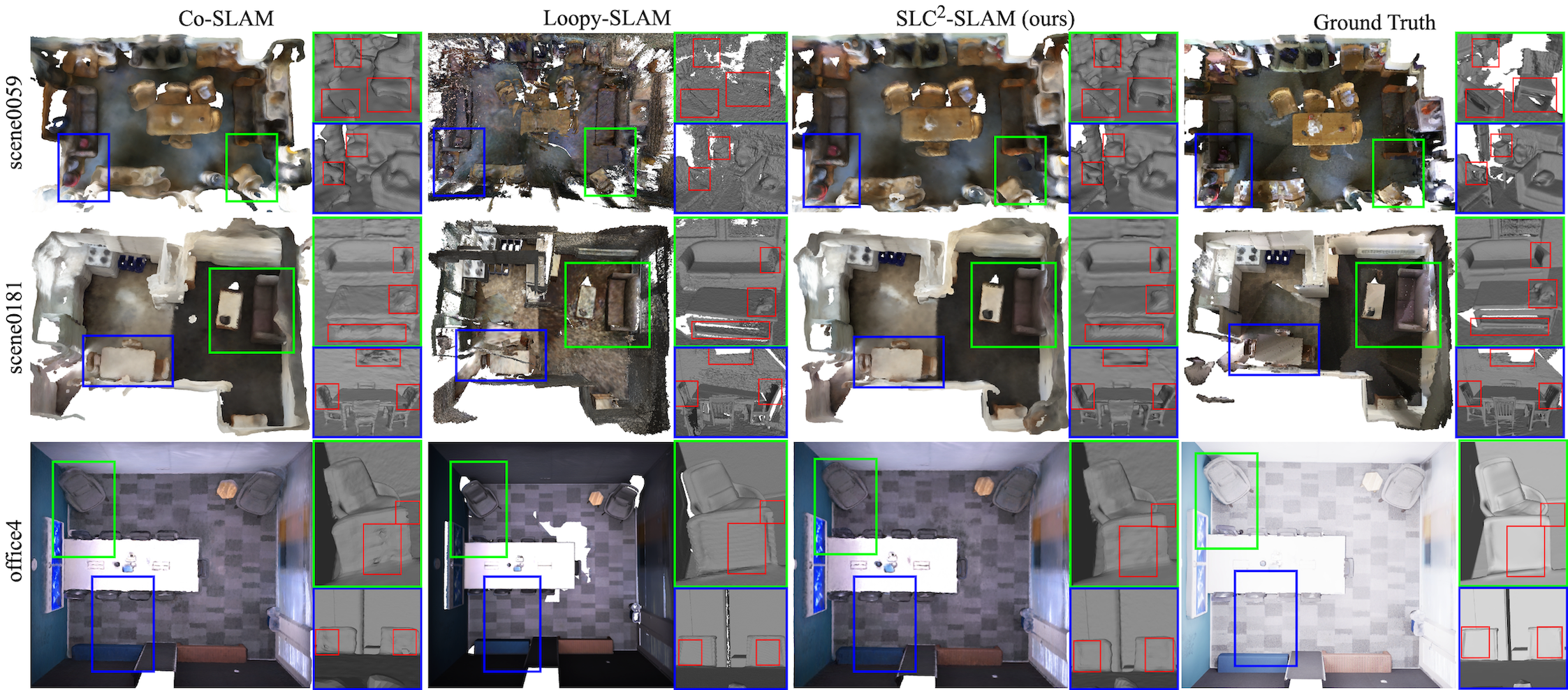}
    \vspace{-3ex}
    \caption{{Reconstruction examples of Co-SLAM~\cite{coslam}, Loopy-SLAM~\cite{loopyslam}, and our SLC$^2$-SLAM on the ScanNet and Replica datasets. 
    Compared to Loopy-SLAM, our reconstructions are more complete for Replica scenes and 
    better aligned and less noisy for ScanNet scenes. 
    Compared to Co-SLAM, ours are  more complete and less noisy for both datasets. Zoomed in views are provided with highlights for better visualization.}
    }
 \label{fig:recon}
 \vspace{-3ex}
\end{figure*}

\textbf{Reconstruction:} The reconstruction results, both quantitative and qualitative, are presented in Table ~\ref{tab:recon-replica} and Fig.~\ref{fig:recon}. While we aimed to compare our system with other loop-closure-enabled methods, only the one by Bruns \textit{et al.}~\cite{bruns2024} provided these metrics. Thus, we compared SLC$^2$-SLAM against three more systems without loop closure, including our baseline Co-SLAM~\cite{coslam}. As shown, our SLC$^2$-SLAM significantly outperforms all other methods across all three metrics, with a considerable margin in each, underscoring the impact of our loop closure on reconstruction quality.

\begin{table*}[t]
\vspace{1ex}

\resizebox{\textwidth}{!}{  
  \begin{minipage}{\textwidth}  
    \centering
    \caption{Reconstruction results on Replica. We mark the best results with \textbf{Bold} and second best with \underline{Underline}}
    \vspace{-2ex}
    \begin{tabular}{@{\extracolsep\fill}llccccccccc}
      \toprule
      Methods & Metric & Room-0 & Room-1 & Room-2 & Office-0 & Office-1 & Office-2 & Office-3 & Office-4 & Avg. \\
      \midrule
    
      
       \multirow{3}{*}{Co-SLAM~\cite{coslam}} 
      & \textbf{Acc.}[cm]$\downarrow$ 
      & \underline{2.11} & \underline{1.68} & \underline{1.99} & \underline{1.57} & \underline{1.31} & \underline{2.84} & \underline{3.06} & \underline{2.23} & \underline{2.10}  \\
      & \textbf{Comp.}[cm]$\downarrow$ 
      & \underline{2.02} & \underline{1.81} & \underline{1.96} & \underline{1.56} &\underline{1.59} & {2.43} & \underline{2.72} & \underline{2.52} & \underline{2.08}  \\
     & \textbf{Comp. Ratio}[$<5$cm \%]$\uparrow$ 
      & \underline{95.26}& \underline{95.19} & \underline{93.58} & \underline{96.09} & {94.65} & 91.63 & \underline{90.72} & \underline{90.44} & \underline{93.44}  \\
      \midrule

      \multirow{3}{*}{Hu \textit{et al.}~\cite{hu2024}} 
      & \textbf{Acc.}[cm]$\downarrow$ 
      & 2.54 & 2.70 & 2.25 & 2.14 & 2.80 & 3.58 & 3.46 & 2.68 & 2.77  \\
      & \textbf{Comp.}[cm]$\downarrow$ 
      & 2.41 & 2.26 & 2.46 & 1.76 & 1.94 & 2.56 & {2.93} & 3.27 & 2.45 \\
      & \textbf{Comp. Ratio}[$<5$cm \%]$\uparrow$ 
      &93.22 &94.75 &93.02 &96.04 & {94.77}& {91.89}& 90.17& 88.46&
       92.79 \\
      \midrule

      \multirow{3}{*}{Bruns \textit{et al.}~\cite{bruns2024}} 
      & \textbf{Acc.}[cm]$\downarrow$ 
      & 2.63 & 2.25 & 2.86 & 1.88 & 2.07 & 3.45 & 4.92 & 2.98 & 2.88  \\
      & \textbf{Comp.}[cm]$\downarrow$ 
      & 2.25 & 1.86 & 3.57 & 1.67 & 1.79 & \underline{2.34} & \textbf{2.69} & 2.67 & 2.36 \\
      & \textbf{Comp. Ratio}[$<5$cm \%]$\uparrow$ 
      & 93.23 & 94.98 & 89.62 & 95.59 &\underline{93.34} & \underline{91.35}& 89.40 & 89.34 & 92.11 \\
      \midrule
      
       \multirow{3}{*}{SLC$^2$-SLAM (ours)} 
      & \textbf{Acc.}[cm]$\downarrow$ 
      & \textbf{1.42} & \textbf{1.31} & \textbf{1.29} & \textbf{1.19} & \textbf{1.09} & \textbf{2.67} & \textbf{2.56} & \textbf{1.54} & \textbf{1.63}  \\
      & \textbf{Comp.}[cm]$\downarrow$ 
      & \textbf{1.41} & \textbf{1.24} & \textbf{1.37} & \textbf{1.16} & \textbf{1.04} & \textbf{1.51} & 3.57 & \textbf{1.57} & \textbf{1.61}  \\
      & \textbf{Comp. Ratio}[$<5$cm \%]$\uparrow$ 
      & \textbf{99.36}& \textbf{99.96} & \textbf{98.58} & \textbf{99.42} & \textbf{99.59} & \textbf{95.85} & \textbf{91.11} & \textbf{98.15} & \textbf{97.75}  \\
      \bottomrule
    \end{tabular}
    \label{tab:recon-replica}
  \end{minipage}
}
\vspace{-1ex}
\end{table*}

\textbf{Memory and Runtime:} 
Our SLC$^2$-SLAM operates efficiently, consuming only 2GB video memory. Although all experiments were conducted on a NVIDIA 4090 GPU, the system can run on any GPU with a minimum of 4GB memory. In contrast, other loop-implemented NeRF SLAM, such as Loopy-SLAM~\cite{loopyslam}, require GPUs with at least 12GB of memory.
For runtime, the tracking and mapping processes of SLC$^2$-SLAM are on par with Co-SLAM~\cite{coslam}. Our loop detection and pose graph optimization achieve, on average, 1.6 seconds and 0.5 seconds per loop, respectively. Comparatively, Loopy-SLAM~\cite{loopyslam} needs 12 seconds.

\section{Conclusion}
In this paper, we present SLC$^2$-SLAM, a NeRF-SLAM system featuring a simple yet highly effective loop closure method. Our approach leverages on-the-fly learned latent codes, originally introduced to assist 3D scene reconstruction, and repurposes them as local features for global descriptor aggregation. To ensure these sampled latent codes accurately represent the current view, we introduce a semantic-guided stratified sampling, drawing on semantic information also decoded from the latent codes. We evaluate our SLC$^2$-SLAM on two publicly available datasets,  comparing it to various NeRF-SLAM systems, both with and without loop closure, and demonstrate its superior performance.

\bibliographystyle{ieeetr}
\bibliography{root}

\end{document}